\documentclass[letterpaper]{article} 
\usepackage{aaai2026} 
\nocopyright
\usepackage{times} 
\usepackage{helvet} 
\usepackage{courier} 
\usepackage[hyphens]{url} 
\usepackage{graphicx} 
\usepackage{natbib} 
\usepackage{caption} 

\usepackage{amsmath}
\usepackage{amssymb}
\usepackage{booktabs}
\usepackage{multirow}
\usepackage{algorithm}
\usepackage{algpseudocode}
\usepackage{subcaption}
\usepackage{makecell}
\usepackage{listings}
\usepackage{tabularx}
\usepackage[T1]{fontenc}
\usepackage[utf8]{inputenc}

\title{Time Series Augmented Generation for Financial Applications}

\author{Anton Kolonin, Alexey Glushchenko, Evgeny Bochkov, Abhishek Saxena \\ SingularityNET Foundation \\ \texttt{akolonin@gmail.com}
}

\begin{document}
\maketitle
\begin{abstract}
Evaluating the reasoning capabilities of Large Language Models (LLMs) for complex, quantitative financial tasks is a critical and unsolved challenge. Standard benchmarks often fail to isolate an agent's core ability to parse queries and orchestrate computations. To address this, \textbf{we introduce a novel evaluation methodology and benchmark designed to rigorously measure an LLM agent's reasoning for financial time-series analysis.} We apply this methodology in a large-scale empirical study using our framework, Time Series Augmented Generation (TSAG), where an LLM agent delegates quantitative tasks to verifiable, external tools. Our benchmark, consisting of 100 financial questions, is used to compare multiple SOTA agents (e.g., GPT-4o, Llama 3, Qwen2) on metrics assessing tool selection accuracy, faithfulness, and hallucination. The results demonstrate that capable agents can achieve near-perfect tool-use accuracy with minimal hallucination, validating the tool-augmented paradigm. \textbf{Our primary contribution is this evaluation framework and the corresponding empirical insights into agent performance, which we release publicly to foster standardized research on reliable financial AI.}
\end{abstract}

\section{Introduction}
\label{sec:introduction}

The financial domain demands timely and accurate insights derived from complex, dynamic data \citep{tsay2005analysis}. Natural language offers an intuitive interface for accessing this information, yet effectively bridging conversational queries with the necessary underlying quantitative analysis remains a significant hurdle. Large Language Models (LLMs) have demonstrated remarkable capabilities in natural language understanding and generation \citep{brown2020language, wu2023bloomberggpt}. However, their application to finance is often constrained by limitations in precise numerical computation, robust temporal reasoning, and reliable grounding in volatile, high-frequency financial time series data \citep{Blasco2024UncertaintySurvey}. Directly applying LLMs to tasks requiring high analytical fidelity can lead to inaccurate or unsubstantiated outputs. Conversely, traditional quantitative models and libraries \citep{boxjenkins1970}, while accurate, lack accessible natural language interfaces, limiting their usability for broader audiences.

This capability gap hinders the seamless integration of sophisticated data analysis into interactive financial workflows. Standard Retrieval-Augmented Generation (RAG) systems primarily retrieve textual documents \citep{lewis2020retrieval} and are insufficient for queries demanding real-time computation over numerical time series. Recent RAG adaptations for time series forecasting focus on retrieving similar historical data sequences \citep{wu2025enhancing_FinSeer}, a different task from executing diverse, on-demand analytical computations required for complex financial question-answering (Q\&A) systems.

While tool-use is a promising solution, its application in finance demands extreme reliability. A key obstacle is the lack of standardized methods for evaluating an LLM agent's core reasoning capabilities for quantitative tasks. It is crucial to distinguish between failures in agentic reasoning (e.g., choosing the wrong tool) and failures in tool execution (e.g., noisy live data). \textbf{This paper tackles this evaluation challenge directly.} We introduce a methodology and benchmark designed to isolate and rigorously measure an agent's ability to parse financial queries, select appropriate computational tools, and extract correct parameters.

We conduct our analysis using Time Series Augmented Generation (TSAG), a framework we developed as a testbed for this purpose. TSAG operates as a \textbf{Tool-Augmented RAG} system.

\textbf{Our main contributions are primarily methodological and empirical:}
\begin{itemize}
    \item \textbf{A novel evaluation methodology and benchmark} for isolating and measuring an LLM agent's core reasoning skills (parsing, tool selection, parameter extraction) for financial Q\&A, independent of live data noise.
    \item \textbf{The first comprehensive empirical study} comparing a wide range of modern LLM agents (Llama 3.x, Qwen2, GPT-4o variants) for this specific task, revealing key trade-offs in their reliability, accuracy, and performance.
    \item \textbf{The public release of our benchmark and evaluation framework} as open-source artifacts to foster standardized and reproducible research into reliable financial AI agents.
\end{itemize}

This paper presents the related work (Section \ref{sec:related_work}), describes the TSAG framework (Section \ref{sec:methodology}), experimental setup (Section \ref{sec:experiments}) and results (Section \ref{sec:results}), and derives the conclusion (Section \ref{sec:conclusion}).

\section{Related Work}
\label{sec:related_work}

Our work integrates insights from four key areas: LLMs in finance, time series analysis, Retrieval-Augmented Generation (RAG), and tool-using agents.

While LLMs like BloombergGPT \citep{wu2023bloomberggpt} excel at financial NLP tasks such as sentiment analysis \citep{loughran2011liability} and summarization \citep{jin2020hooks}, they often fail at the precise numerical grounding required for quantitative queries, motivating our tool-based approach. Established financial time series methods, from classical models like ARIMA \citep{boxjenkins1970} and GARCH \citep{engle1982autoregressive} to deep learning approaches \citep{hochreiter1997lstm, vaswani2017attention, lim2021time}, are powerful for forecasting but lack native natural language interfaces for diverse Q\&A. The need for verifiable computation is further highlighted by challenges in model uncertainty \citep{Blasco2024UncertaintySurvey}.

Standard RAG retrieves text \citep{lewis2020retrieval}, and recent adaptations for time series focus on improving forecasting by retrieving similar historical data sequences \citep{wu2025enhancing_FinSeer}. Our TSAG framework differs significantly by employing an agent to invoke computational \textbf{tools} for analytical Q\&A, rather than retrieving data for prediction. This aligns our work with the growing paradigm of tool-using LLM agents \citep{Schick2023Toolformer, Chase2022LangChain, Patil2023Gorilla}, a recognized direction for time series analysis \citep{zhang2023large_ts_survey}. While other agentic systems for time series focus on integrating external text \citep{ma2024news_to_forecast} or using complex multi-agent loops for forecasting \citep{liu2025timecap, liu2025timexl}, TSAG is distinct. It focuses specifically on financial Q\&A, using a single agent to orchestrate a curated set of \textbf{predefined, reliable quantitative functions} to ensure verifiable accuracy.

\section{Methodology: The TSAG Framework}
\label{sec:methodology}

TSAG operates as a Tool-Augmented RAG system using an LLM agent to orchestrate specialized tools. \textbf{For this study, we employ a controlled experimental design to isolate and rigorously evaluate the LLM agent's core reasoning capabilities.} This approach allows us to measure an agent's performance on parsing, tool selection, and parameter extraction, distinct from confounding variables like network latency or live data noise that would arise from end-to-end execution.

\begin{figure}[t]
    \centering
    \includegraphics[width=\columnwidth]{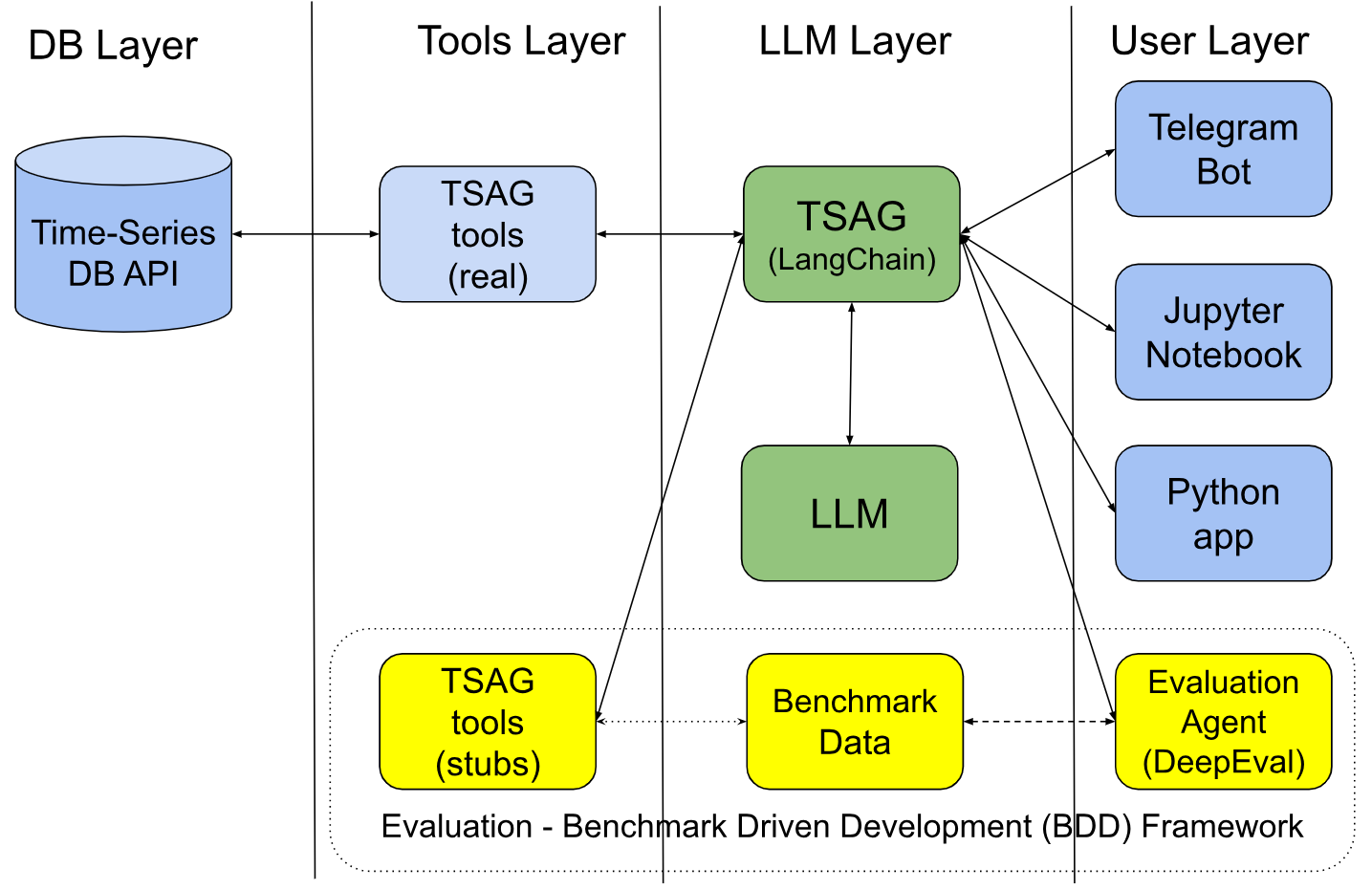} 
    \caption{The TSAG Tool-Augmented RAG architecture and workflow.}
    \label{fig:architecture}
\end{figure}

\subsection{Architecture} \label{subsec:architecture}
The architecture consists of four layers, presented in Figure \ref{fig:architecture}. 
First, there is a User Layer with "front-end" components which may be represented by a Telegram bot, Jupyter Notebook for research purposes, any end-user application (e.g. in Python) or the Evaluation Agent as part of the evaluation framework based on DeepEval \cite{DeepEval2023} being presented in this paper. Second, the LLM Layer consists of the TSAG kernel based on LangChain \cite{Chase2022LangChain}, connecting to one of the selected LLM agents - either self-hosted locally or cloud-hosted such as ChatGPT or DeepSeek. Third, the Tools Layer contains the tools plugged into the TSAG kernel as a "grounding" functions. Finally, the time series (TS) database layer (DB Layer) provides API for accessing the TS data. 

For the evaluation purposes discussed in this paper, the Tools Layer contains grounding function "stubs" with expected responses hard-coded according to the Benchmark Data. The latter makes it possible to abstract from specific transient temporal data in the TS DB to perform \textbf{Benchmark-Driven Development (BDD)} of the TSAG framework on the basis of the Benchmark Data, supporting designated set of grounding functions.

\subsection{Workflow} \label{subsec:workflow}
Given a natural language query (NLQ), TSAG proceeds via LLM-orchestrated steps to obtain a natural language response (NLR):
\begin{enumerate}
    \item \textbf{Query Parsing \& Tool Selection:} LLM agent parses NLQ, identifies intent, extracts parameters (using defaults from subsection \ref{subsec:defaults}) and selects tool(s) (subsection \ref{subsec:tools}) based on prompt descriptions (subsection \ref{subsec:llm_prompting}.
    \item \textbf{Tool Execution:} System invokes selected tool(s) (e.g., \texttt{volatility(...)} according to \citet{Parkinson1980}) via predefined Python code, querying data sources and performing verifiable TS computation.
    \item \textbf{Result Synthesis:} Tool returns structured result (e.g., \verb|{'volatility_percent': 5.0}|).
    \item \textbf{Response Generation:} LLM agent synthesizes result into a grounded NLR answering the NLQ.
\end{enumerate}
This structured workflow ensures calculations are handled by dedicated, verifiable logic.

\subsection{Grounding Functions (Tools)} \label{subsec:tools}
A core TSAG component is a library of specialized Python functions serving as tools. This proof-of-concept (POC) focuses on cryptocurrency trading data sourced from a Time-Series Database, as shown in Figure \ref{fig:architecture}. We implement tools based on the NLQ $\rightarrow$ Function $\rightarrow$ NLR structure. \textbf{As our primary goal is to benchmark the agent's reasoning, our evaluation deliberately utilizes tool "stubs" with predefined outputs.} This methodological choice ensures that our metrics reflect the agent's ability to correctly invoke the tool, rather than the computational correctness of the tool itself on dynamic data. The initial tool set, prioritizing reliability for this study, covers the following:
\begin{itemize}
    \item \textbf{Seasonality/Pattern Analysis:} Tools (\verb|peak_traded_volume|, \verb|lowest_traded_volume|, \verb|round_the_clock_pattern|, \verb|abnormal_deviations|) identifying recurring/anomalous volume patterns via statistical analysis.
    \item \textbf{Price and Volatility Analysis:} Tools (\verb|price|, \verb|volatility|) retrieving current data or calculating historical metrics (e.g., Parkinson volatility \citep{Parkinson1980}) along with respective prediction tools (\verb|predict_price|, \verb|predict_volatility|).
    \item \textbf{Correlation Analysis:} Tools (\verb|correlation_between_exchanges|, \verb|correlation_between_tokens|) computing Pearson correlation coefficients \citep{Pearson1895} across different instruments and exchanges.
    \item \textbf{Metadata Retrieval:} Tools (\verb|get_base_tokens|, \verb|get_exchanges|, \verb|get_valid_time_units|, etc.) that provide valid parameters for the tools listed above.
\end{itemize}

The grounding functions used in the TSAG POC are implemented in Python. These tools encapsulate specific analytical logic corresponding to common financial queries according to business specification in Table \ref{tab:tools_summary_appendix}. The stubs of the functions delivered as part of the benchmark presented in this work are containing specific expected responses hard-coded in \verb|tsag/tools.py| source code according to the benchmark in \verb|tests/benchmark.tsv|.

Future work includes implementing more sophisticated "real" functions and expanding the scope of applications to traditional finance.

\begin{table*}[htbp]
    \centering
    \small 
    \setlength{\tabcolsep}{4pt}
    \begin{tabular*}{\textwidth}{@{\extracolsep{\fill}} p{4.7cm} p{3.9cm} p{5.8cm} l @{}}
    \toprule
    \textbf{Function Name} & \textbf{Description} & \textbf{Key Parameters (Defaults)} & \textbf{Returns} \\
    \midrule
    \texttt{peak\_traded\_volume} & Finds times (e.g., day of week) with highest volume within periods, exceeding relative threshold. &
      \texttt{base\_token}: str \newline
      \texttt{quote\_token}: str (default='USDT') \newline
      \texttt{exchange}: str (default='BINANCE') \newline
      \texttt{time\_interval}: int (default=1) \newline
      \texttt{time\_unit}: str (default='year') \newline
      \texttt{period\_unit}: str (default='week') \newline
      \texttt{granularity\_unit}: str (default='day') \newline
      \texttt{threshold\_percent}: float (default=5.0)
      & list[str] \\
    \midrule
    \texttt{lowest\_traded\_volume} & Finds times with lowest volume within periods, below relative threshold. &
      \texttt{base\_token}: str \newline
      \texttt{quote\_token}: str (default='USDT') \newline
      \texttt{exchange}: str (default='BINANCE') \newline
      \texttt{time\_interval}: int (default=1) \newline
      \texttt{time\_unit}: str (default='year') \newline
      \texttt{period\_unit}: str (default='week') \newline
      \texttt{granularity\_unit}: str (default='day') \newline
      \texttt{threshold\_percent}: float (default=5.0)
      & list[str] \\
    \midrule
    \texttt{round\_the\_clock\_pattern} & Combines peak and lowest volume times for pattern summary. &
      \texttt{base\_token}: str \newline
      \texttt{quote\_token}: str (default='USDT') \newline
      \texttt{exchange}: str (default='BINANCE') \newline
      \texttt{time\_interval}: int (default=1) \newline
      \texttt{time\_unit}: str (default='year') \newline
      \texttt{period\_unit}: str (default='week') \newline
      \texttt{granularity\_unit}: str (default='day') \newline
      \texttt{threshold\_percent}: float (default=5.0)
      & tuple \\
    \midrule
    \texttt{abnormal\_deviations} & Finds recent time points with volume deviations exceeding historical norms by threshold. &
      \texttt{base\_token}: str \newline
      \texttt{quote\_token}: str (default='USDT') \newline
      \texttt{exchange}: str (default='BINANCE') \newline
      \texttt{time\_interval}: int (default=1) \newline
      \texttt{time\_unit}: str (default='year') \newline
      \texttt{period\_unit}: str (default='week') \newline
      \texttt{granularity\_unit}: str (default='day') \newline
      \texttt{threshold\_percent}: float (default=5.0)
      & tuple \\
    \midrule
    \texttt{price} & Gets the latest price within the lookback window. &
      \texttt{base\_token}: str \newline
      \texttt{quote\_token}: str (default='USDT') \newline
      \texttt{exchange}: str (default='BINANCE') \newline
      \texttt{time\_interval}: int (default=1) \newline
      \texttt{time\_unit}: str (default='day')
      & float \\
    \midrule
    \texttt{volatility} & Calculates historical price volatility (Parkinson method) over window. &
      \texttt{base\_token}: str \newline
      \texttt{quote\_token}: str (default='USDT') \newline
      \texttt{exchange}: str (default='BINANCE') \newline
      \texttt{time\_interval}: int (default=1) \newline
      \texttt{time\_unit}: str (default='day')
      & float \\
    \midrule
    \texttt{predict\_price}* & Predicts price for the next window (POC: simple extrapolation). &
      \texttt{base\_token}: str \newline
      \texttt{quote\_token}: str (default='USDT') \newline
      \texttt{exchange}: str (default='BINANCE') \newline
      \texttt{time\_interval}: int (default=1) \newline
      \texttt{time\_unit}: str (default='day')
      & float \\
    \midrule
    \texttt{predict\_volatility}* & Predicts volatility for next window (POC: simple extrapolation). &
      \texttt{base\_token}: str \newline
      \texttt{quote\_token}: str (default='USDT') \newline
      \texttt{exchange}: str (default='BINANCE') \newline
      \texttt{time\_interval}: int (default=1) \newline
      \texttt{time\_unit}: str (default='day')
      & float \\
    \midrule
    \texttt{correlation\_between\_tokens} & Computes Pearson price correlation between two tokens on one exchange. &
      \texttt{base\_token\_a}: str \newline
      \texttt{base\_token\_b}: str \newline
      \texttt{quote\_token}: str (default='USDT') \newline
      \texttt{exchange}: str (default='BINANCE') \newline
      \texttt{time\_interval}: int (default=7) \newline
      \texttt{time\_unit}: str (default='day')
      & float \\
    \bottomrule
    \end{tabular*}
    \caption{Summary of key grounding functions (tools) in TSAG POC. Parameters are listed with type and default value. *Prediction tools use simple extrapolation.}
    \label{tab:tools_summary_appendix}
\end{table*}

\subsection{Function Parameters and Defaults} \label{subsec:defaults}
The LLM extracts parameters from the NLQ. Predefined defaults (e.g., \verb|quote_token='USDT'|, \verb|exchange='BINANCE'|) handle short queries with poorly defined input parameters.

\subsection{LLM Agents} \label{subsec:llm_prompting}
We evaluated several LLMs (Table \ref{tab:main_results}) as agents, selected to explore trade-offs between size, cost, accuracy, hallucinations and run-time performance. Initial experiments involved use of different prompts, however at some point we have sorted out that using standard LangChain contextualization together with sufficient context size does not require specific prompts in order to maximize accuracy and minimize hallucinations. In particular, we discovered that context of 8192 is sufficient for holding the tool context for our set of tools, while smaller contexts of 4096 and default 2048 is insufficient. Specifically, insufficient context does not fit all tools to have their parameters (function arguments) identified properly. 

\section{Experimental Setup} \label{sec:experiments}
We evaluated TSAG reliability, accuracy, level of hallucinations, and run-time efficiency across different LLM agents.

\subsection{Tasks}
We run the evaluation framework against benchmark consisting of 100 natural language questions with different levels of brevity corresponding to the tools described above in subsection \ref{subsec:tools}: seasonality and patterns, price/volatility, correlation, and metadata retrieval.

\subsection{Benchmark and Test Corpus}
\label{subsec:benchmark}
The test corpus used for evaluation consists of 100 items each represented by triplet derived from the original business specification, including the original natural language query (NLQ, as a "zero shot" sample), a set of expected words or numbers to be found in the text of the natural language response (NLR), and the full expected NLR text in the expected wording at the expected level of detail.

\subsection{Evaluation Framework and Metrics} \label{subsec:evaluation_framework}
We evaluated across LLM agents using the evaluation framework based on DeepEval \cite{DeepEval2023}, having the following metrics evaluated based on the benchmark, having the DeepEval backed up with locally hosted Qwen2 7B LLM agent \citep{bai2024qwen2}.
\begin{itemize}
    \item \textbf{Return Rate (RR):} For each of the benchmark items - end-to-end execution success indicator, set to 1.0 if any non-empty text result is returned by the TSAG framework in response to the benchmark NLQ, and 0.0 otherwise. Across the entire benchmark - the average result, the higher the better.
    \item \textbf{Match Accuracy (MA):} \textbf{This metric programmatically validates the agent's reasoning by checking if the generated tool name and all extracted parameters are an exact match with a ground-truth object.} It is a strict, non-subjective measure of the agent's ability to correctly invoke a tool.
    \item \textbf{LLM-accessed Accuracy (LA):} For each of the benchmark items - measure of accuracy comparing the expected NLR from benchmark against the actual NLR, assessed by \texttt{DeepEval} \citep{DeepEval2023} in the range between 0.0 and 1.0. Across the entire benchmark - the average result, the higher the better.
    \item \textbf{Hallucination Rate (HR):} For each of the benchmark items - degree of contextual divergence of the actual NLR text away from the expected one, assessed by \texttt{DeepEval} in the range between 0.0 and 1.0. Across the entire benchmark - the average result, the lower the better.
    \item \textbf{Seconds per Query (SPQ):} For each of the benchmark items - the amount of seconds to obtain NLR given NLQ. Across the entire benchmark - the average result, the lower the better.
\end{itemize}

\subsection{Comparative Analysis} \label{subsec:comparative_analysis}
Our primary analysis compares the performance of different LLM agents, hosted locally or available online in the cloud, given the TSAG framework (Table \ref{tab:main_results}) to identify models best suited for orchestrating financial tools, a key goal of our work. We evaluated these LLM agents against our benchmark. As expected, raw models not augmented with the tools provided ultimate Return Rate but showed zero Match Accuracy and negligible LLM-assessed Accuracy with high Hallucination Rate, confirming TSAG's necessity for grounded quantitative Q\&A. These baseline results are omitted from Table \ref{tab:main_results} for clarity.

\subsection{Implementation Details \& Hyper-parameters} \label{subsec:implementation}
For TSAG development and evaluation we used Python 3.11, langchain 0.3.20, langchain-core 0.3.45, langchain-deepseek-official 0.1.0, langchain-ollama 0.2.2, langchain-openai 0.3.8, deepeval 2.5.4. 
LLM agents hosted locally were accessed via Ollama v0.1.32 for local models and official vendor APIs (OpenAI, DeepSeek) accessed in April 2025. The following models were evaluated.
\begin{itemize}
    \item Llama 3.1 (8B) \citep{meta:llama3.1:ollama:8b}
    \item Llama 3.2 (3.2B) \citep{MetaLlama3.2_Ref}
    \item Qwen2 (0.5B, 1.5B, 7B) \citep{bai2024qwen2}
    \item Qwen2.5 (0.5B, 1.5B, 3B, 7B) \citep{Lu2024Qwen2.5}
    \item GPT-4o \citep{OpenAI_GPT4o}
    \item GPT-4o-mini \citep{OpenAI_GPT4omini}
    \item DeepSeek-V3 (API) \citep{DeepSeekV3_Ref}
\end{itemize}

Some LLM agents that we were considering to use initially, including local Gemma 7B \citep{gemma2024}, DeepSeek hosted online \citep{DeepSeekV3_Ref}, and online Qwen \citep{Lu2024Qwen2.5} were excluded from evaluation results due to incompatibility with the required tooling in the used version of LangChain \citep{10690817}.

The following LLM parameters and hyper-parameters were used, overriding the defaults for respective LLM agents.

\begin{itemize}
    \item LLM temperatures (\verb|temperature|): [0.0, 1.0]
    \item Random seeds (\verb|seed|): [1, 10, 100]
    \item Length of LLM context window in tokens (\verb|num_ctx|): 8192
    \item The maximum number of tokens LLM is allowed to generate (\verb|num\_predict|): 512
    \item The maximum number of retries for GPT and DeepSeek LLM agents (\verb|max\_retries|): 2
    \item The maximum number of retries for all LLM agents, in case if no text is generated (\verb|retries|): 5
\end{itemize}

The search of the parameters and hyper-parameters was performed incrementally along with development course. Lower numbers or defaults of hyper-parameters were chosen at the beginning of the study and then we were incrementally increasing them in the course of debugging and benchmarking till the return rate and accuracy metrics reached the plateau and stopped improving further.

\begin{figure}[t]
    \centering
    \includegraphics[width=\columnwidth]{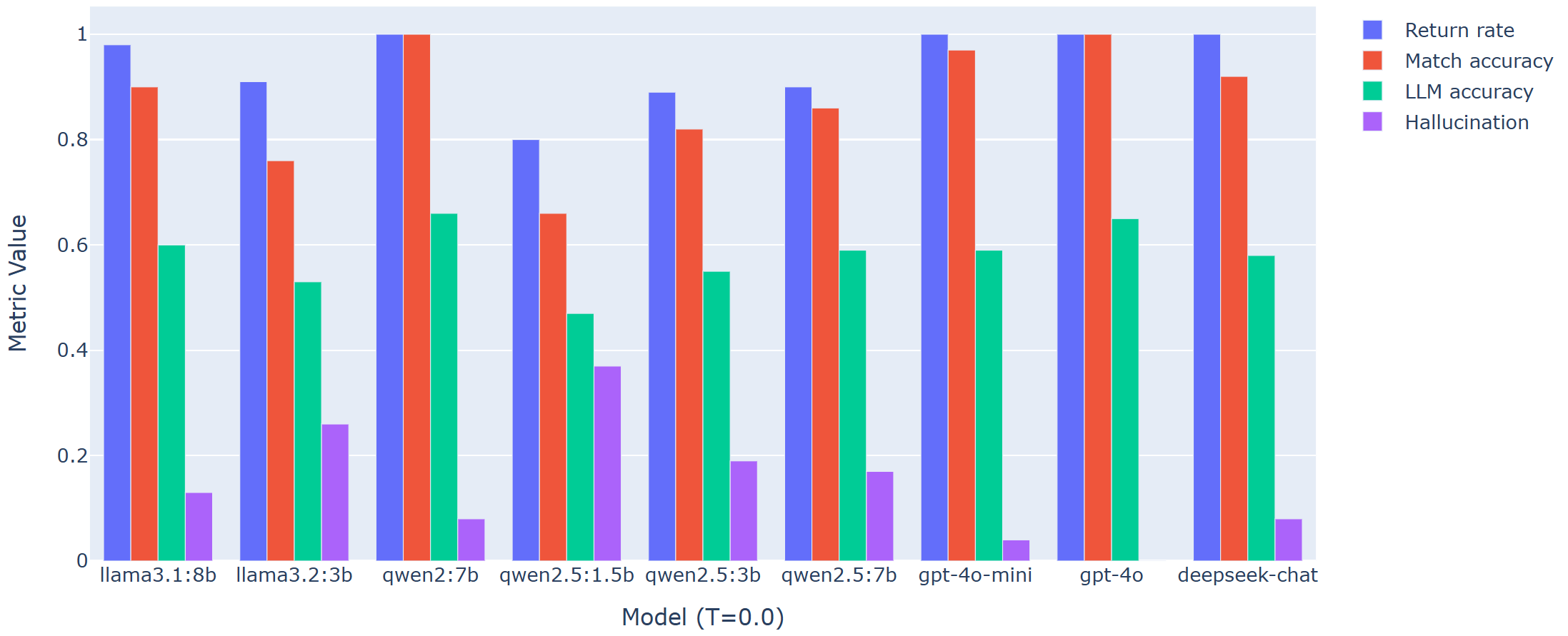}
    \caption{TSAG performance with different LLM agents (Temperature = 0.0). Metrics: return rate, match accuracy, LLM-measured accuracy and hallucination rate measured by DeepEval framework. Figure~\ref{fig:results_plots} shows results for Temperature = 1.0.}
    \label{fig:results_t0}
\end{figure}

\begin{figure*}[htbp]
    \centering
    \begin{subfigure}[b]{0.49\textwidth}
        \centering
        \includegraphics[width=\linewidth]{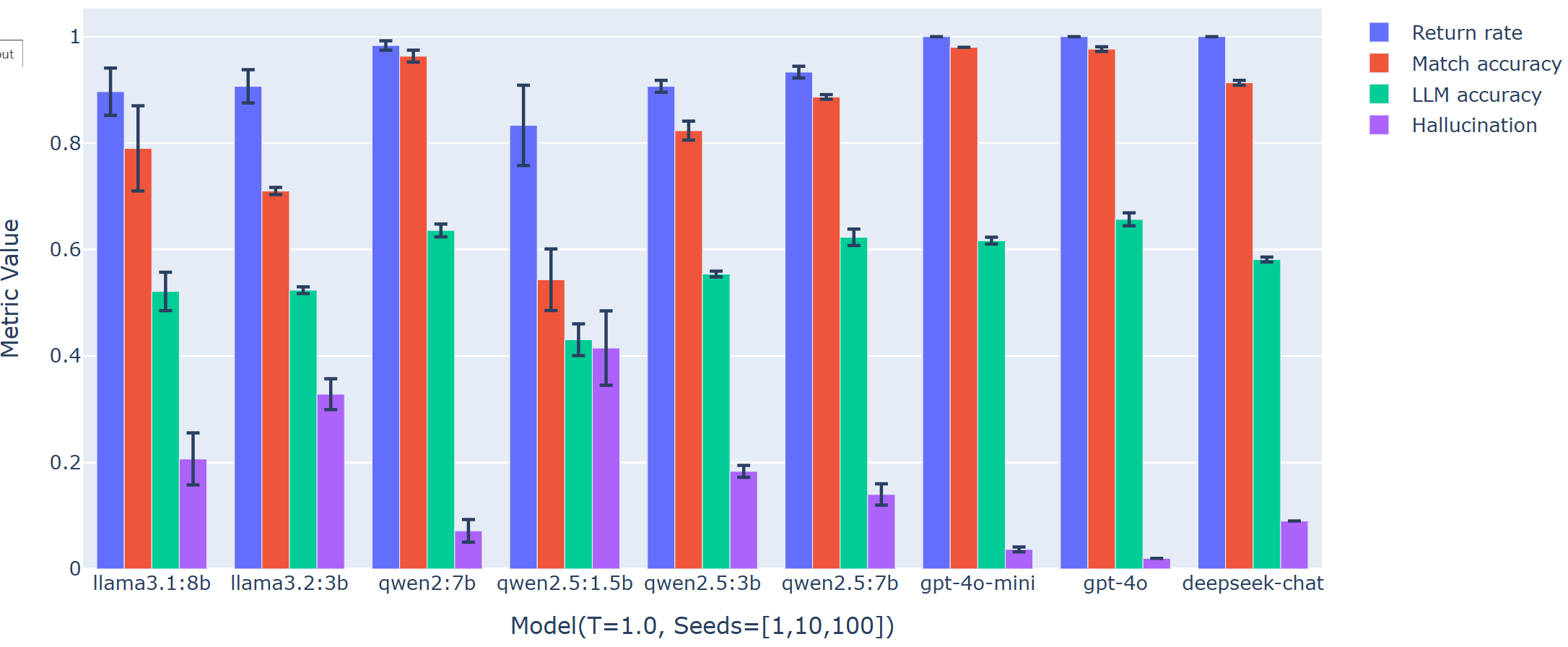}
        \caption{Metrics: return rate, match accuracy, LLM-measured accuracy and hallucination rate measured by DeepEval framework.}
        \label{fig:results_metrics}
    \end{subfigure}
    \hfill
    \begin{subfigure}[b]{0.49\textwidth}
        \centering
        \includegraphics[width=\linewidth]{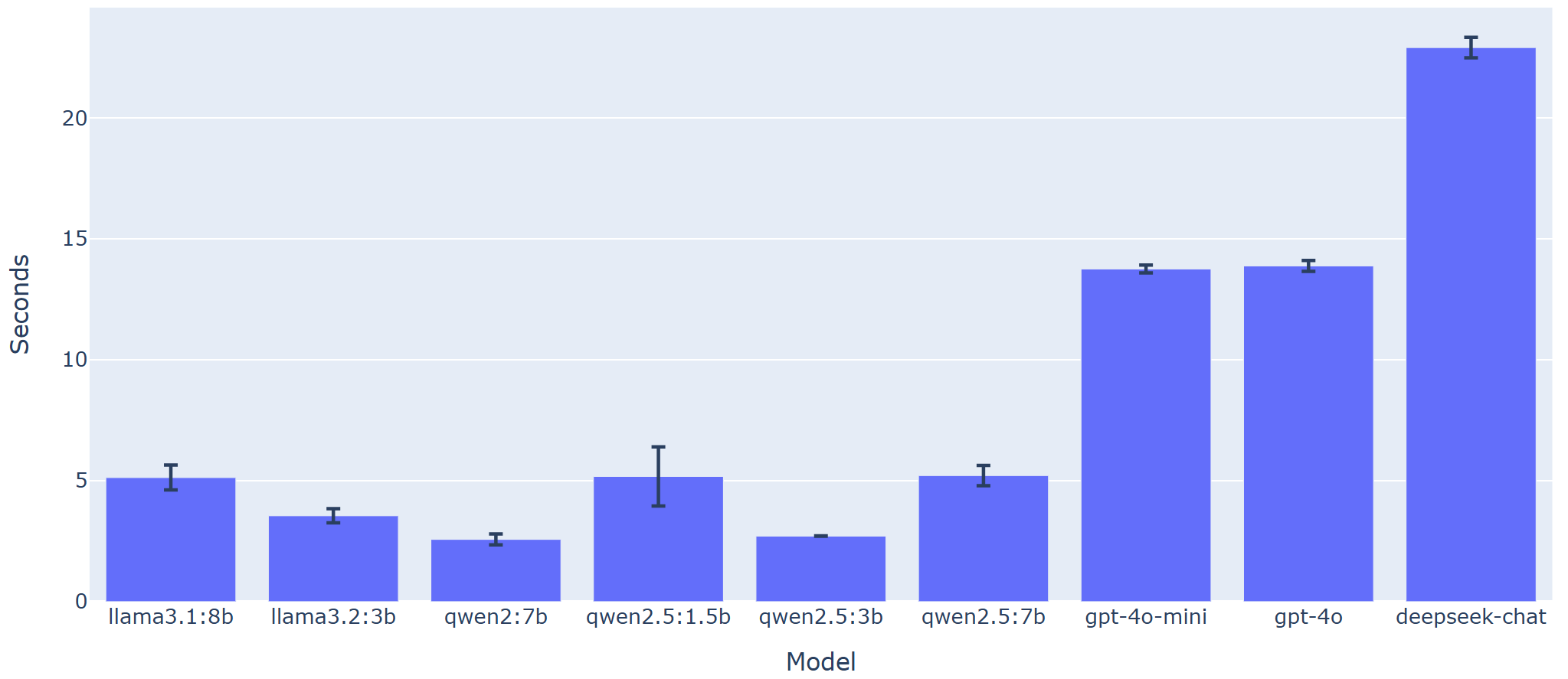}
        \caption{Seconds spent by query (SPQ). Use of GPT and especially DeepSeek takes longer, compared to locally hosted models.}
        \label{fig:results_spq}
    \end{subfigure}
    \caption{TSAG performance with different LLM agents as average over 3 runs (Temperature = 1.0). Error bars indicate run variability as mean percentage error (MPE). Figure~\ref{fig:results_t0} shows results for Temperature = 0.0.}
    \label{fig:results_plots}
\end{figure*}

We evaluated the LLM agents with different temperatures: Temperature=0.0 was used for them most "conservative" responses (Figure \ref{fig:results_t0}), Temperature=1.0 was used for the greatest "diversity". In case of Temperature=1.0, we used 3 different random seeds [1, 10, 100] for 3 runs (Table \ref{tab:main_results}, Figure \ref{fig:results_plots}).

The following hardware was used for the evaluation: MSI Raider GE77HX 12UGS notebook with 12th Gen Intel(R) Core(TM) i7-12800HX 2.00 GHz, 32.0 GB RAM, 23.9 GB GPU NVIDIA GeForce RTX 3070 Ti Laptop GPU. The computational budget in hours was taking about 2 hours for each run of the benchmark including TSAG with LLM agent execution and evaluation carried our with DeepEval. The total research time with all experimental runs and debugging was about 3 machine-months.

\begin{table}[t]
\centering\small
\begin{tabular}{lrrrrrr} \toprule
\textbf{LLM agent} & \textbf{RR}$\uparrow$ & \textbf{MA}$\uparrow$ & \textbf{LA}$\uparrow$ & \textbf{HR}$\downarrow$ & \textbf{SPQ}$\downarrow$ \\
\midrule
\makecell[l]{Llama 3.1 (8B) \\ \citep{meta:llama3.1:ollama:8b}} & 0.98 & 0.90 & 0.60 & 0.13 & 5 \\
\makecell[l]{Llama 3.2 (3.2B) \\ \citep{MetaLlama3.2_Ref}} & 0.91 & 0.76 & 0.53 & 0.26 & 3 \\
\makecell[l]{Qwen2 (7B) \\ \citep{bai2024qwen2}} & \textbf{1.00} & \textbf{1.00} & \textbf{0.66} & \textbf{0.08} & \textbf{2} \\
\makecell[l]{Qwen2.5 (1.5B) \\ \citep{Lu2024Qwen2.5}} & 0.80 & 0.66 & 0.47 & 0.37 & 6 \\
\makecell[l]{Qwen2.5 (3B) \\ \citep{Lu2024Qwen2.5}} & 0.89 & 0.82 & 0.55 & 0.19 & 3 \\
\makecell[l]{Qwen2.5 (7B) \\ \citep{Lu2024Qwen2.5}} & 0.90 & 0.86 & 0.59 & 0.17 & 5 \\
\makecell[l]{GPT-4o (API) \\ \citep{OpenAI_GPT4o}} & \textbf{1.00} & \textbf{1.00} & \textbf{0.65} & \textbf{0.02} & \textbf{2} \\
\makecell[l]{GPT-4o-mini (API) \\ \citep{OpenAI_GPT4omini}} & 1.00 & 0.97 & 0.59 & 0.04 & 3 \\
\makecell[l]{DeepSeek-V3 (API) \\ \citep{DeepSeekV3_Ref}} & 1.00 & 0.92 & 0.58 & 0.08 & 14 \\
\bottomrule
\end{tabular}
\caption{TSAG performance with different LLM agents (Temperature = 0.0). Metrics: return rate (RR), match accuracy (MA), LLM-measured accuracy (LA), hallucination rate (HR), and seconds per query (SPQ). Arrows indicate preferred direction. Best results are bold.}
\label{tab:main_results}
\end{table}

\section{Results and Discussion} \label{sec:results}
We analyzed the performance of TSAG, focusing on comparing LLM agents, using multiple runs with different random seeds with non-zero temperature to assess the reliability of our evaluations.

\subsection{Quantitative Results} \label{subsec:quantitative_results}
Our quantitative evaluation confirms the viability of the TSAG framework, with most agents achieving high Return Rates, indicating successful end-to-end execution (Table \ref{tab:main_results}). \textbf{Crucially, the results highlight a clear performance distinction based on the agent's underlying capabilities.}

\textbf{The top-performing agents, GPT-4o and Qwen2 (7B), demonstrated flawless tool orchestration, achieving a perfect Match Accuracy of 1.00.} In terms of response quality, these models also excelled: Qwen2 (7B) reached the highest LLM-assessed Accuracy at \textbf{0.66}, while GPT-4o achieved a remarkably low Hallucination Rate of just \textbf{0.02}. This confirms that with a capable agent, the TSAG framework can produce highly reliable and factually grounded responses.

\textbf{In contrast, smaller models struggled significantly.} For instance, Llama 3.2 (3.2B) and Qwen2.5 (1.5B) achieved much lower Match Accuracies of only 0.76 and 0.66, respectively, with correspondingly higher Hallucination Rates. This suggests that the complex reasoning required for robust tool use is an emergent capability of larger models. The variability across runs, indicated by MPE error bars (Figure \ref{fig:results_plots}), further supports this, showing higher consistency for top-performing agents.

Comparing the deterministic "conservative" performance at Temperature=0.0 shown in Figure \ref{fig:results_t0} with the "diverse" (which can be considered "exploratory") results at Temperature=1.0 shown in Figure \ref{fig:results_metrics} shows that the top-performing models deterministically maintain high Return Rate, Match Accuracy, LLM-assessed Accuracy, and low Hallucination Rate. Zero temperature generally leads to slight reductions in hallucination across the board, reinforcing its suitability for high-stakes applications requiring maximal factuality.

Latency analysis (Figure \ref{fig:results_spq}) reveals significant differences. Local models like Qwen2 7B (2.2s) and the smaller Qwen2.5 3B (2.8s) offer fast responses. API models vary, with GPT-4o-mini (~2.4-3s) being relatively quick, while DeepSeek-V3 exhibited high latency (~14s) in our tests. This highlights a crucial trade-off between accessing potentially highest-performing proprietary models versus leveraging efficient open models for lower latency.

\begin{figure*}[t]
    \centering
    \includegraphics[width=\textwidth]{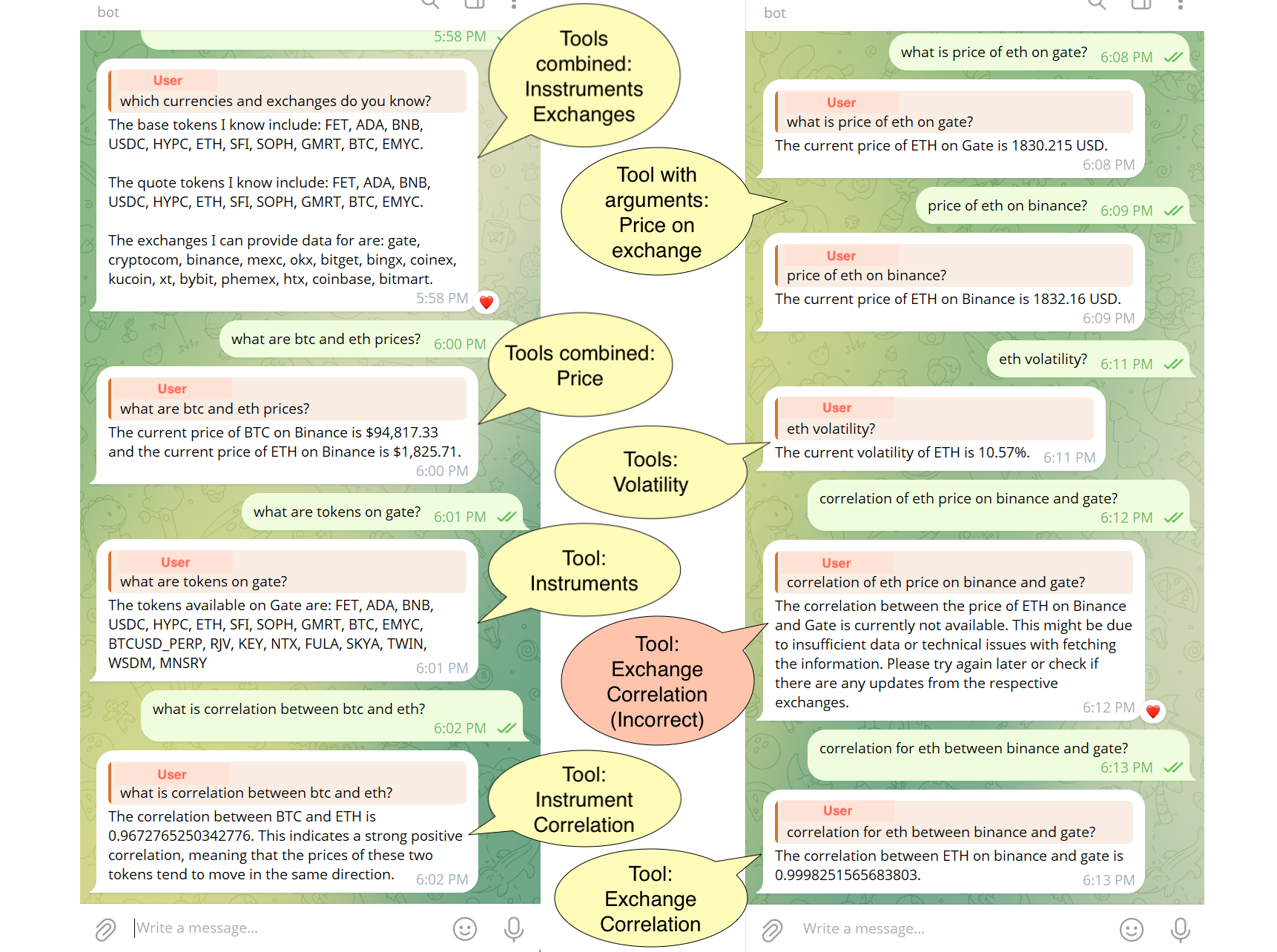}
    \caption{An example of the TSAG framework implemented as a Telegram chatbot, showing successful tool selection for various queries and one instance of an incorrect tool choice (Exchange Correlation instead of Instrument Correlation).}
    \label{fig_bot_annotation}
\end{figure*}

\subsection{Qualitative Analysis} \label{subsec:qualitative_analysis}
The TSAG framework presented above has been implemented through a chat bot which interact with real time time-series data, having the tool functions implemented as "stubs" in the evaluation framework replace with real quantitative analysis functions for price and volume analysis, correlation, prediction and pattern mining like presented in Figure~\ref{fig_bot_annotation}. From the Figure~\ref{fig_bot_annotation} it is seen that most of the queries are interpreted correctly, picking the right tool functions, identifying input arguments for them appropriately and even combining calls to multiple different tools or the same tool with different arguments correctly, except on case on the figure.

Manual review of generated responses complements the quantitative findings. Successful cases demonstrate the LLM agent's ability to parse complex natural language, correctly identify and parameterize the appropriate tool (e.g., handling multiple constraints in a query handled by function like \verb|peak_traded_volume(...)|), and synthesize the numerical or list-based output into a fluent, accurate sentence in natural language. For example, given "What was the price correlation between BTC and ETH quoted in USDT on BINANCE exchange in the past 7 days?", a high-performing agent correctly invokes \verb|correlation_between_tokens| with extracted parameters and generates a response like "The price correlation between tokens BTC and ETH quoted in USDT on BINANCE exchange in the past 7 days trading window is 1.0." Common failure modes, particularly with less capable agents or ambiguous queries, included selecting an incorrect tool, failing to extract all necessary parameters (relying incorrectly on defaults), or minor hallucinations where the synthesized text slightly misrepresents the exact nuance of the tool's output (though major fabrications were rare with top models, reflected in low HR scores).

\subsection{Discussion} \label{subsec:discussion}
The experimental results strongly validate Time-Series Augmented Generation (TSAG) approach for enabling reliable quantitative Q\&A over financial time series using LLMs. By delegating computation to predefined, verifiable tools, TSAG effectively grounds LLM responses, achieving high accuracy scores (Table \ref{tab:main_results}) and mitigating the inherent weaknesses of LLMs in complex numerical reasoning \citep{Blasco2024UncertaintySurvey}. The comparison across different agents underscores the critical role of the LLM's underlying capabilities; models like GPT-4o and Qwen2 (7B) excel at the multi-step reasoning involved in parsing, tool selection/invocation, and synthesis. The performance drop-off with smaller models suggests that robust instruction-following and parameter extraction are non-trivial requirements for successful tool orchestration in this domain.

Our analysis of run variability (MPE error bars, Figure \ref{fig:results_plots}) indicates good stability for the top-performing models across different random seeds with high temperature, lending confidence to their reliability. The comparison between Temperature=0.0 (Figure \ref{fig:results_t0}) and Temperature=1.0 average results (Figure \ref{fig:results_metrics}) suggests that while deterministic generation slightly reduces hallucination, performance on core accuracy metrics remains high for capable agents across these temperatures, offering flexibility depending on application needs (consistency vs. minor linguistic variation).

The TSAG approach, using predefined tools, offers significant advantages in terms of reliability and interpretability compared to alternatives considered in our design phase, such as agents generating Python code on the fly to create and execute plans \citep{Yao2022ReAct, Suris2023ViperGPT} or end-to-end fine-tuning of a Large Language-Numeric Model (LLNM). While less flexible than code generation, the verifiable nature of the tools is paramount in the high-stakes financial domain. The challenges encountered, primarily related to natural language ambiguity and tool coverage, highlight key areas for future work. Improving the TSAG robustness through advanced prompting, agent fine-tuning, or incorporating query clarification steps could enhance performance. Expanding the tool library, potentially incorporating more sophisticated forecasting models beyond the current POC stubs, and developing mechanisms for multi-tool composition are necessary steps to broaden TSAG's analytical capabilities and address more complex financial queries. The latency differences observed also motivate further exploration of optimized local models like Qwen2 (7B).

\section{Conclusion} \label{sec:conclusion}
In this paper, we presented and evaluated a Time-Series Augmented Generation (TSAG) approach, implemented as a Tool-Augmented RAG framework designed to empower LLM agents with reliable quantitative analysis capabilities for financial time series Q\&A. By orchestrating calls to a library of specialized, predefined computational tools, TSAG effectively grounds language model outputs in verifiable data analysis, addressing critical limitations of LLMs in numerical reasoning and factuality within the demanding financial domain.

Our comprehensive experiments, comparing multiple state-of-the-art LLM agents (Llama 3.x, Qwen2 variants, GPT-4o variants, DeepSeek-V3), demonstrate the viability and effectiveness of this approach. We show that capable agents like GPT-4o and Qwen2 7B, when integrated into TSAG, achieve high accuracy and low hallucination in executing the tool-based pipeline, accurately matching tool outputs, generating faithful responses with minimal hallucination (validated by DeepEval). We analyzed performance trade-offs across model size, accuracy (RR, MA, LA) and hallucination (HR) metrics, latency (SPQ), and consistency (MPE), providing insights into agent selection. Furthermore, analysis of deterministic (Temperature=0.0) versus stochastic (Temperature=1.0) generation highlights the framework's stability and potential for tuning based on application requirements for factuality versus linguistic variation.

We provide a benchmark that can be used to evaluate other LLM agents to improve the described evaluation metrics, and can also be extended to obtain richer and more functional sets of tools that extend the applicability of TSAG. 
This work validates tool-augmentation as a powerful and pragmatic strategy... Key future directions include expanding the tool library and, critically, \textbf{enhancing the agent's ability to handle complex multi-step reasoning by exploring agentic architectures like ReAct.} Furthermore, we will address robustness by \textbf{incorporating Chain-of-Thought (CoT) prompting for better explainability and by developing a paraphrased version of our benchmark} to test for linguistic variations, while continuing to prioritize ethical considerations and responsible AI development.

\section{Limitations} \label{sec:limitations}
Our work has limitations:
\begin{itemize}
    \item \textbf{Tool Coverage \& Brittleness:} Scope limited by predefined tools; cannot answer out-of-scope queries. Performance of the non-stub tools depends on their implementation and content of the actual time-series database. Tools may be brittle to API/data format changes.
    \item \textbf{Tool Correctness:} Accuracy hinges on meticulous tool implementation and valid data sources. Further work on uncertainty quantification \citep{Blasco2024UncertaintySurvey} is also needed.
    \item \textbf{Natural Language Robustness:} \textbf{Our benchmark, while comprehensive, does not systematically test for robustness against linguistic variations such as paraphrasing or intentionally ambiguous phrasing. The brittleness of the agent to such changes is an important, unmeasured variable in this work.}
    \item \textbf{Static Tools:} No dynamic code generation, limiting flexibility compared to some agent approaches. Reliance on LangChain-compatible tooling excluded some models (e.g., online Qwen).
    \item \textbf{Evaluation Scope:} Focused on crypto-finance Q\&A domain; generalization to traditional finance requires significant tool adaptation. Run variability analysis based on 3 seeds provides initial insights but more runs would be beneficial.
    \item \textbf{Compositionality:} \textbf{The current single-step workflow of TSAG cannot handle complex queries requiring multi-tool reasoning chains (e.g., 'Find assets with high volatility and positive recent news'). This is a significant limitation for advanced financial analysis.}
    \item \textbf{Sensitivity to Agent/Parameters:} Performance varies significantly with LLM choice and temperature, necessitating careful tuning and selection.
    \item \textbf{Baseline Scope:} Our comparative analysis focuses on general-purpose LLM agents. We did not include benchmarks against finance-specialized LLMs operating without tools, which could further highlight the necessity of the tool-use paradigm.
\end{itemize}

\section{Ethical Considerations} \label{sec:ethics}
Deploying and using TSAG, especially in production environment, requires addressing ethical concerns, as follows.
\begin{itemize}
    \item \textbf{Potential Risks} Improper use of any financial instrument can cause financial damage to a user. From this perspective, grounding an LLM agent performance with manually predefined function with control over hallucination level, as we do in our work, can mitigate the risk. Anyway, using our TSAG framework in production, as using any financial instrument on the market requires personal responsibility and awareness.       
    \item \textbf{Personally Identifying Info Or Offensive Content:} The benchmark we created do not include any personally identifying info or offensive content.
    \item \textbf{Transparency vs. Opacity:} Tool use enhances computational transparency. LLM reasoning remains partially opaque; we provide code details with tool stubs, TSAG framework and evaluation framework.
    \item \textbf{Accuracy and Reliability:} Crucial in finance. Tool accuracy dependence requires rigorous testing. Low hallucination rates are confirmed in our study for top performing models. Users must understand outputs are tool-based information, not infallible financial advice.
    \item \textbf{Data Bias:} In production use, reliance on historical time-series data may reflect historical market biases (e.g., asset popularity, exchange-specific patterns). Auditing for bias propagation is needed.
    \item \textbf{Potential Misuse:} Generating convincing analyses requires safeguards against misinformation/manipulation. Not a substitute for professional advice.
    \item \textbf{Data Privacy:} Uses public data; adaptation for private data requires robust security/privacy protocols.
    \item \textbf{Reproducibility:} Enhanced by releasing code for function stubs with tool descriptions, entire TSAG framework and evaluation framework. API access and specific model versions may limit full replication.
    \item \textbf{Fairness \& Equity:} When using the TSAG platform for production purposes, potential biases may arise if the data or actual implementation of real tools favors certain assets or exchanges. Expanding tool coverage requires attention to equitable representation.
\end{itemize}


\bibliography{references}

\makeatletter
\@ifundefined{isChecklistMainFile}{
  \newif\ifreproStandalone
  \reproStandalonetrue
}{
  \newif\ifreproStandalone
  \reproStandalonefalse
}
\makeatother

\ifreproStandalone
\documentclass[letterpaper]{article}
\usepackage[submission]{aaai2026}
\setlength{\pdfpagewidth}{8.5in}
\setlength{\pdfpageheight}{11in}
\usepackage{times}
\usepackage{helvet}
\usepackage{courier}
\usepackage{xcolor}
\frenchspacing

\begin{document}
\fi
\setlength{\leftmargini}{20pt}
\makeatletter\def\@listi{\leftmargin\leftmargini \topsep .5em \parsep .5em \itemsep .5em}
\def\@listii{\leftmargin\leftmarginii \labelwidth\leftmarginii \advance\labelwidth-\labelsep \topsep .4em \parsep .4em \itemsep .4em}
\def\@listiii{\leftmargin\leftmarginiii \labelwidth\leftmarginiii \advance\labelwidth-\labelsep \topsep .4em \parsep .4em \itemsep .4em}\makeatother

\setcounter{secnumdepth}{0}
\renewcommand\thesubsection{\arabic{subsection}}
\renewcommand\labelenumi{\thesubsection.\arabic{enumi}}

\newcounter{checksubsection}
\newcounter{checkitem}[checksubsection]

\newcommand{\checksubsection}[1]{%
  \refstepcounter{checksubsection}%
  \paragraph{\arabic{checksubsection}. #1}%
  \setcounter{checkitem}{0}%
}

\newcommand{\checkitem}{%
  \refstepcounter{checkitem}%
  \item[\arabic{checksubsection}.\arabic{checkitem}.]%
}
\newcommand{\question}[2]{\normalcolor\checkitem #1 #2 \color{blue}}
\newcommand{\ifyespoints}[1]{\makebox[0pt][l]{\hspace{-15pt}\normalcolor #1}}

\section*{Reproducibility Checklist}
\vspace{1em}
\hrule
\vspace{1em}

\checksubsection{General Paper Structure}
\begin{itemize}

\question{Includes a conceptual outline and/or pseudocode description of AI methods introduced}{(yes/partial/no/NA)}
yes

\question{Clearly delineates statements that are opinions, hypothesis, and speculation from objective facts and results}{(yes/no)}
yes

\question{Provides well-marked pedagogical references for less-familiar readers to gain background necessary to replicate the paper}{(yes/no)}
yes

\end{itemize}
\checksubsection{Theoretical Contributions}
\begin{itemize}

\question{Does this paper make theoretical contributions?}{(yes/no)}
no

	\ifyespoints{\vspace{1.2em}If yes, please address the following points:}
        \begin{itemize}
	
	\question{All assumptions and restrictions are stated clearly and formally}{(yes/partial/no)}
	NA

	\question{All novel claims are stated formally (e.g., in theorem statements)}{(yes/partial/no)}
	NA

	\question{Proofs of all novel claims are included}{(yes/partial/no)}
	NA

	\question{Proof sketches or intuitions are given for complex and/or novel results}{(yes/partial/no)}
	NA

	\question{Appropriate citations to theoretical tools used are given}{(yes/partial/no)}
	NA

	\question{All theoretical claims are demonstrated empirically to hold}{(yes/partial/no/NA)}
	NA

	\question{All experimental code used to eliminate or disprove claims is included}{(yes/no/NA)}
	NA
	
	\end{itemize}
\end{itemize}

\checksubsection{Dataset Usage}
\begin{itemize}

\question{Does this paper rely on one or more datasets?}{(yes/no)}
yes

\ifyespoints{If yes, please address the following points:}
\begin{itemize}

	\question{A motivation is given for why the experiments are conducted on the selected datasets}{(yes/partial/no/NA)}
	yes

	\question{All novel datasets introduced in this paper are included in a data appendix}{(yes/partial/no/NA)}
	yes

	\question{All novel datasets introduced in this paper will be made publicly available upon publication of the paper with a license that allows free usage for research purposes}{(yes/partial/no/NA)}
	yes

	\question{All datasets drawn from the existing literature (potentially including authors' own previously published work) are accompanied by appropriate citations}{(yes/no/NA)}
	NA

	\question{All datasets drawn from the existing literature (potentially including authors' own previously published work) are publicly available}{(yes/partial/no/NA)}
	NA

	\question{All datasets that are not publicly available are described in detail, with explanation why publicly available alternatives are not scientifically satisficing}{(yes/partial/no/NA)}
	NA

\end{itemize}
\end{itemize}

\checksubsection{Computational Experiments}
\begin{itemize}

\question{Does this paper include computational experiments?}{(yes/no)}
yes

\ifyespoints{If yes, please address the following points:}
\begin{itemize}

	\question{This paper states the number and range of values tried per (hyper-) parameter during development of the paper, along with the criterion used for selecting the final parameter setting}{(yes/partial/no/NA)}
	yes

	\question{Any code required for pre-processing data is included in the appendix}{(yes/partial/no)}
	NA

	\question{All source code required for conducting and analyzing the experiments is included in a code appendix}{(yes/partial/no)}
	yes

	\question{All source code required for conducting and analyzing the experiments will be made publicly available upon publication of the paper with a license that allows free usage for research purposes}{(yes/partial/no)}
	yes
        
	\question{All source code implementing new methods have comments detailing the implementation, with references to the paper where each step comes from}{(yes/partial/no)}
	no

	\question{If an algorithm depends on randomness, then the method used for setting seeds is described in a way sufficient to allow replication of results}{(yes/partial/no/NA)}
	yes

	\question{This paper specifies the computing infrastructure used for running experiments (hardware and software), including GPU/CPU models; amount of memory; operating system; names and versions of relevant software libraries and frameworks}{(yes/partial/no)}
	yes

	\question{This paper formally describes evaluation metrics used and explains the motivation for choosing these metrics}{(yes/partial/no)}
	yes

	\question{This paper states the number of algorithm runs used to compute each reported result}{(yes/no)}
	yes

	\question{Analysis of experiments goes beyond single-dimensional summaries of performance (e.g., average; median) to include measures of variation, confidence, or other distributional information}{(yes/no)}
	yes

	\question{The significance of any improvement or decrease in performance is judged using appropriate statistical tests (e.g., Wilcoxon signed-rank)}{(yes/partial/no)}
	yes

	\question{This paper lists all final (hyper-)parameters used for each model/algorithm in the paper’s experiments}{(yes/partial/no/NA)}
	yes

\end{itemize}
\end{itemize}
\ifreproStandalone
\end{document}
\fi

\appendix

\end{document}